\documentclass[11pt]{article}

\usepackage[preprint]{acl}

\usepackage{times}
\usepackage{latexsym}

\usepackage[T1]{fontenc}

\usepackage[utf8]{inputenc}

\usepackage{microtype}

\usepackage{inconsolata}

\usepackage{graphicx}

\title{Teaching and Critiquing Conceptualization and Operationalization in NLP}

\author{Vagrant Gautam \\
  Heidelberg Institute for Theoretical Studies \\
  Heidelberg, Germany \\
  \texttt{vagrant.gautam@h-its.org} \\}

\begin{document}
\maketitle
\begin{abstract}
NLP researchers regularly invoke abstract concepts like ``interpretability,'' ``bias,'' ``reasoning,''  and ``stereotypes,'' without defining them.
Each subfield has a shared understanding or conceptualization of what these terms mean and how we should treat them, and this shared understanding is the basis on which operational decisions are made:
Datasets are built to evaluate these concepts, metrics are proposed to quantify them, and claims are made about systems. 
But what do they mean, what \textit{should} they mean, and how should we measure them?
I outline a seminar I created for students to explore these questions of conceptualization and operationalization, with an interdisciplinary reading list and an emphasis on discussion and critique.
\end{abstract}

\section{Introduction}

Natural language processing and computational linguistics are relatively young, with the main professional organization (ACL), flagship conference (ACL), and the CL journal only being founded in the 1960s and 70s.
As the field matures, we have entered a period of critical reflection about practices and definitions.
For example, since the question of ethical NLP was raised \citep{hovy-spruit-2016-social}, we now have dedicated workshops \citep{ws-2017-acl}, policies,\footnote{https://ethics.aclweb.org/} tutorials \citep{benotti-etal-2025-navigating}, and teaching \citep{friedrich-zesch-2021-crash} for it.

More recently, NLP researchers have begun to interrogate how we define and measure abstract concepts such as bias and understanding, revealing that proposed methods for measuring or mitigating bias are ``poorly matched to their motivation'' \citep{blodgett-etal-2020-language}, and that current benchmarking practices in natural language understanding ``obscure the abilities that we want our benchmarks to measure'' \citep{bowman-dahl-2021-will}.
Understanding these conceptual debates and using them to move the field forward are a fundamental skill for training NLP scholars rather than NLP engineers \citep{prasad-davis-2024-training-nlp}.
This is why I developed a seminar on conceptualization and operationalization in NLP, with the following learning objectives:

\begin{itemize}
    \setlength\itemsep{0em}
    \item Read and critique papers (both interdisciplinary and more conventional NLP papers)
    \item Critically evaluate aspects of conceptualization (defining an abstract concept) and operationalization (creating empirical measures of it, such as datasets and metrics) in NLP
    \item Design NLP projects in ways that address critiques and push the discipline forward
\end{itemize}

\section{Structure}

The seminar was offered to Masters students and advanced Bachelors students in computational linguistics and computer science.
After some introductory sessions, we cycled through different concepts every 1-2 weeks, for which everyone was expected to read and discuss four papers:
A \textit{critique/conceptual} paper about the concept, to use as a lens to critique three other recent \textit{content} papers in NLP that use that concept (see Section \ref{sec:assigned-readings} for details on the readings).
Students presented and led discussions about each concept in pairs.
To hone students' skills at critiquing research, I follow principes of scaffolded learning \citep{wood-bruner-ross-1976-role}, by assigning papers that demonstrate the skill, and giving targeted feedback on assignments.

\paragraph{Introductory sessions.}
I used the first three sessions to establish expectations and  course logistics, to teach reading, conceptualization and operationalization, and to give a sample presentation/discussion.
Students chose their preferred concepts and dates, which informed the schedule and content for the rest of the class.
As reading papers is part of the hidden curriculum \citep{wilson-2024-documenting-unwritten}, I felt it was important to teach it explicitly with: \citet{keshav-2007-how,eisner-2009-how,carey-et-al-2020-ten,fruehwald-2022-reading}.
Then we discussed definitions of conceptualization \citep{subramonian-etal-2023-takes}, operationalization \citep{steidl-werum-2019-operationalization}, and how they impact research as well as society.
Finally, I presented the concept of \textit{names} and led a discussion about it as a sample for the synthesis and critique I expected from students.

\paragraph{Course requirements.}
I used \textit{assignments} with discussion prompts to 
encourage reading and thinking about the papers before student-led \textit{presentation/discussions} about each concept, during which I also graded \textit{engagement} from other students.
Students could also write an optional \textit{final report} designing a novel NLP project or re-imagining one of the content papers we read in a way that addressed our critique.
See Appendix \ref{app:course-requirements} for details.

\section{Readings}
\label{sec:assigned-readings}

A full list of concepts, critique papers and content papers is shown in Appendix \ref{app:assigned-readings}.
For future iterations of the course, content papers can be swapped out for more recent or relevant work (e.g., the hype around DeepSeek's release made \citet{deepseekai2025deepseekr1incentivizingreasoningcapability} an obvious late addition to the \textit{reasoning} readings).
Importantly, the critique/conceptual papers, which are often interdisciplinary, hold up regardless of current NLP trends.

For \textit{interpretability}, we read perspectives from a machine learning researcher \citep{lipton-2018-mythos} and a philosopher \citep{Krishnan2020-KRIAIA-3}, that critique interpretability as being a slippery concept and even unnecessary.
\citeauthor{Krishnan2020-KRIAIA-3}'s \citeyearpar{Krishnan2020-KRIAIA-3} distinction between causal and justificatory explanations grounded our readings of the popular and more technical papers, Patchscopes \citep{ghandeharioun2024patchscopes} and RAVEL \citep{huang-etal-2024-ravel}.

As for \textit{explainability}, we read \citet{MILLER20191}, which summarizes findings from philosophy, cognitive science, and psychology that are relevant to human explanations and thus useful for explainability in NLP.
We also read 'Attention is not Explanation' \citep{jain-wallace-2019-attention} and a response to it titled 'Attention is not not Explanation' \citep{wiegreffe-pinter-2019-attention}, which is itself a critique.
Beyond the scientific content of these papers, reading them is  also an important part of understanding the history of the field and important debates.
Another such example is \citet{bender-koller-2020-climbing} and \citet{michael-2020-dissect}, which has prompted rich discussion in other pedagogical settings as well \citep{guerzhoy-2024-occams-razor}.

\section{Encouraging Discussion and Critique}

Direction and feedback were centrally important for students to develop the skill of critique.
In their assignments, students were asked to synthesize commonalities and differences across papers and point out what authors did especially well, or any choices they were skeptical about.
Some students had trouble with this at the beginning, instead just summarizing what each individual paper did, but targeted feedback and examples helped.
By the end of the term, some students were even bringing up points covered in other literature that we had not read in this course, and I could send them paper recommendations rather than constructive feedback in my responses to their assignments.

Assignments were due before the sessions where we discussed the concept, so discussions were also high quality.
I emphasized that all of our perspectives are important to co-create knowledge in the classroom \citep{mccrae-2024-co-creational}, and I explicitly graded presenters on their classroom management and whether they gave everyone an opportunity to participate, since discussion-heavy seminars can be challenging for some students.
A third of student evaluations explicitly mentioned this as a positive.
Offering a variety of topics and types of papers also allowed students in this ``mixed'' undergraduate-graduate and interdisciplinary setting to participate, as others have observed \citep{fosler-lussier-2008-strategies}.

The diversity of students in terms of sociodemographic factors like gender and country of origin were also particularly interesting for discussions of sociodemographic concepts such as \textit{bias} \citep{blodgett-etal-2020-language}, \textit{stereotypes} \citep{blodgett-etal-2021-stereotyping}, and \textit{names} \citep{gautam-etal-2024-stop}.
Students also made several connections across this cluster of topics.
Finally, students practised turning critique into action through final project proposals designed to address critique and push the discipline forward.


\section{Conclusion}
I have presented a seminar designed to teach critical thinking about abstract concepts in NLP, with attention to both its social and technical aspects.
Through readings, assignments, presentations, discussions, and final reports, students gained an appreciation for how decisions about conceptualization and operationalization can impact all aspects of our research, as well as public perceptions of NLP technology and tools.

\section*{Limitations}

The primary limitation of this course is that it does not scale in its current form; a small classroom is essential for equal participation and for quality feedback with just one instructor.
Additionally, it is challenging to balance a cohesive reading schedule (e.g., putting related topics such as \textit{names}, \textit{stereotypes}, and \textit{bias} close together) with optimizing for student-preferred presentation dates.
Finally, this course works best when students engage intellectually with the content themselves, rather than outsourcing their thinking to LLMs \citep{guest_2025_17786243}.
This cannot be ensured, but only emphasized by the instructor and encouraged by creating a classroom environment that acknowledges and embraces friction as part of the learning process.

\section*{Ethics Statement}

I designed this class from a standpoint of seeing NLP technologies as fundamentally sociotechnical \citep{selbst-et-al-2019,dhole-2023-large}, value-laden \citep{birhane-et-al-2022-values}, and situated rather than objective \citep{haraway-1988-situated,waseem2021disembodiedmachinelearningillusion}.
I wanted to share this with NLP and CL students who are rarely given the opportunity to engage with such perspectives from science and technology studies (STS) and human-computer interaction (HCI) research, but may nevertheless go on to have power as system designers in academia and industry \citep{scheuerman-et-al-2024-products}.
My pedagogy is informed by critical, feminist and engaged pedagogies \citep{hooks-1994-teaching}, which are important to create a learning environment that feels fundamentally safe to be present in and share, and to broaden epistemological engagement beyond traditional ways of knowing in computer science and STEM \citep{raji-et-al-2021-you}.
This is particularly important when discussing concepts such as race, class, gender, and so on, where \textit{power} is an important consideration that often goes unexamined.

\bibliography{custom}

@inproceedings{mccrae-2024-co-creational,
    title = "Co-Creational Teaching of Natural Language Processing",
    author = "McCrae, John",
    editor = {Al-azzawi, Sana  and
      Biester, Laura  and
      Kov{\'a}cs, Gy{\"o}rgy  and
      Marasovi{\'c}, Ana  and
      Mathur, Leena  and
      Mieskes, Margot  and
      Weissweiler, Leonie},
    booktitle = "Proceedings of the Sixth Workshop on Teaching NLP",
    month = aug,
    year = "2024",
    address = "Bangkok, Thailand",
    publisher = "Association for Computational Linguistics",
    url = "https://aclanthology.org/2024.teachingnlp-1.5/",
    pages = "33--42"
}

@inproceedings{prasad-davis-2024-training-nlp,
    title = "Training an {NLP} Scholar at a Small Liberal Arts College: A Backwards Designed Course Proposal",
    author = "Prasad, Grusha  and
      Davis, Forrest",
    editor = {Al-azzawi, Sana  and
      Biester, Laura  and
      Kov{\'a}cs, Gy{\"o}rgy  and
      Marasovi{\'c}, Ana  and
      Mathur, Leena  and
      Mieskes, Margot  and
      Weissweiler, Leonie},
    booktitle = "Proceedings of the Sixth Workshop on Teaching NLP",
    month = aug,
    year = "2024",
    address = "Bangkok, Thailand",
    publisher = "Association for Computational Linguistics",
    url = "https://aclanthology.org/2024.teachingnlp-1.16/",
    pages = "105--118"
}

@inproceedings{guerzhoy-2024-occams-razor,
    title = "Occam{'}s Razor and Bender and Koller{'}s Octopus",
    author = "Guerzhoy, Michael",
    editor = {Al-azzawi, Sana  and
      Biester, Laura  and
      Kov{\'a}cs, Gy{\"o}rgy  and
      Marasovi{\'c}, Ana  and
      Mathur, Leena  and
      Mieskes, Margot  and
      Weissweiler, Leonie},
    booktitle = "Proceedings of the Sixth Workshop on Teaching NLP",
    month = aug,
    year = "2024",
    address = "Bangkok, Thailand",
    publisher = "Association for Computational Linguistics",
    url = "https://aclanthology.org/2024.teachingnlp-1.18/",
    pages = "128--129"
}

@inproceedings{wilson-2024-documenting-unwritten,
    title = "Documenting the Unwritten Curriculum of Student Research",
    author = "Wilson, Shomir",
    editor = {Al-azzawi, Sana  and
      Biester, Laura  and
      Kov{\'a}cs, Gy{\"o}rgy  and
      Marasovi{\'c}, Ana  and
      Mathur, Leena  and
      Mieskes, Margot  and
      Weissweiler, Leonie},
    booktitle = "Proceedings of the Sixth Workshop on Teaching NLP",
    month = aug,
    year = "2024",
    address = "Bangkok, Thailand",
    publisher = "Association for Computational Linguistics",
    url = "https://aclanthology.org/2024.teachingnlp-1.1/",
    pages = "1--3"
}

@inproceedings{fosler-lussier-2008-strategies,
    title = "Strategies for Teaching ``Mixed'' Computational Linguistics Classes",
    author = "Fosler-Lussier, Eric",
    editor = "Palmer, Martha  and
      Brew, Chris  and
      Xia, Fei",
    booktitle = "Proceedings of the Third Workshop on Issues in Teaching Computational Linguistics",
    month = jun,
    year = "2008",
    address = "Columbus, Ohio",
    publisher = "Association for Computational Linguistics",
    url = "https://aclanthology.org/W08-0205/",
    pages = "36--44"
}

@proceedings{ws-2017-acl,
    title = "Proceedings of the First {ACL} Workshop on Ethics in Natural Language Processing",
    editor = "Hovy, Dirk  and
      Spruit, Shannon  and
      Mitchell, Margaret  and
      Bender, Emily M.  and
      Strube, Michael  and
      Wallach, Hanna",
    month = apr,
    year = "2017",
    address = "Valencia, Spain",
    publisher = "Association for Computational Linguistics",
    url = "https://aclanthology.org/W17-1600/"
}

@inproceedings{hovy-spruit-2016-social,
    title = "The Social Impact of Natural Language Processing",
    author = "Hovy, Dirk  and
      Spruit, Shannon L.",
    editor = "Erk, Katrin  and
      Smith, Noah A.",
    booktitle = "Proceedings of the 54th Annual Meeting of the Association for Computational Linguistics (Volume 2: Short Papers)",
    month = aug,
    year = "2016",
    address = "Berlin, Germany",
    publisher = "Association for Computational Linguistics",
    url = "https://aclanthology.org/P16-2096/",
    doi = "10.18653/v1/P16-2096",
    pages = "591--598"
}

@inproceedings{benotti-etal-2025-navigating,
    title = "Navigating Ethical Challenges in {NLP}: Hands-on strategies for students and researchers",
    author = {Benotti, Luciana  and
      Ducel, Fanny  and
      Fort, Kar{\"e}n  and
      Ivetta, Guido  and
      Jin, Zhijing  and
      Kan, Min-Yen  and
      Lee, Seunghun J.  and
      Li, Minzhi  and
      Mieskes, Margot  and
      Pagano, Adriana},
    editor = "Arase, Yuki  and
      Jurgens, David  and
      Xia, Fei",
    booktitle = "Proceedings of the 63rd Annual Meeting of the Association for Computational Linguistics (Volume 5: Tutorial Abstracts)",
    month = jul,
    year = "2025",
    address = "Vienna, Austria",
    publisher = "Association for Computational Linguistics",
    url = "https://aclanthology.org/2025.acl-tutorials.5/",
    doi = "10.18653/v1/2025.acl-tutorials.5",
    pages = "7--8",
    ISBN = "979-8-89176-255-8"
}

@inproceedings{friedrich-zesch-2021-crash,
    title = "A Crash Course on Ethics for Natural Language Processing",
    author = "Friedrich, Annemarie  and
      Zesch, Torsten",
    editor = "Jurgens, David  and
      Kolhatkar, Varada  and
      Li, Lucy  and
      Mieskes, Margot  and
      Pedersen, Ted",
    booktitle = "Proceedings of the Fifth Workshop on Teaching NLP",
    month = jun,
    year = "2021",
    address = "Online",
    publisher = "Association for Computational Linguistics",
    url = "https://aclanthology.org/2021.teachingnlp-1.6/",
    doi = "10.18653/v1/2021.teachingnlp-1.6",
    pages = "49--51"
}

@inproceedings{bowman-dahl-2021-will,
    title = "What Will it Take to Fix Benchmarking in Natural Language Understanding?",
    author = "Bowman, Samuel R.  and
      Dahl, George",
    editor = "Toutanova, Kristina  and
      Rumshisky, Anna  and
      Zettlemoyer, Luke  and
      Hakkani-Tur, Dilek  and
      Beltagy, Iz  and
      Bethard, Steven  and
      Cotterell, Ryan  and
      Chakraborty, Tanmoy  and
      Zhou, Yichao",
    booktitle = "Proceedings of the 2021 Conference of the North American Chapter of the Association for Computational Linguistics: Human Language Technologies",
    month = jun,
    year = "2021",
    address = "Online",
    publisher = "Association for Computational Linguistics",
    url = "https://aclanthology.org/2021.naacl-main.385/",
    doi = "10.18653/v1/2021.naacl-main.385",
    pages = "4843--4855"
}

@inproceedings{blodgett-etal-2020-language,
    title = "Language (Technology) is Power: A Critical Survey of ``Bias'' in {NLP}",
    author = "Blodgett, Su Lin  and
      Barocas, Solon  and
      Daum{\'e} III, Hal  and
      Wallach, Hanna",
    editor = "Jurafsky, Dan  and
      Chai, Joyce  and
      Schluter, Natalie  and
      Tetreault, Joel",
    booktitle = "Proceedings of the 58th Annual Meeting of the Association for Computational Linguistics",
    month = jul,
    year = "2020",
    address = "Online",
    publisher = "Association for Computational Linguistics",
    url = "https://aclanthology.org/2020.acl-main.485/",
    doi = "10.18653/v1/2020.acl-main.485",
    pages = "5454--5476"
}

@article{keshav-2007-how,
author = {Keshav, S.},
title = {How to read a paper},
year = {2007},
issue_date = {July 2007},
publisher = {Association for Computing Machinery},
address = {New York, NY, USA},
volume = {37},
number = {3},
issn = {0146-4833},
url = {https://doi.org/10.1145/1273445.1273458},
doi = {10.1145/1273445.1273458},
journal = {SIGCOMM Comput. Commun. Rev.},
month = jul,
pages = {83–84},
numpages = {2}
}

@article{carey-et-al-2020-ten,
    doi = {10.1371/journal.pcbi.1008032},
    author = {Carey, Maureen A. AND Steiner, Kevin L. AND Petri, Jr, William A.},
    journal = {PLOS Computational Biology},
    publisher = {Public Library of Science},
    title = {Ten simple rules for reading a scientific paper},
    year = {2020},
    month = {07},
    volume = {16},
    url = {https://doi.org/10.1371/journal.pcbi.1008032},
    pages = {1-6},
    number = {7},

}

@misc{fruehwald-2022-reading,
title={Reading a Technical Paper},
url={https://jofrhwld.github.io/teaching/courses/2022_lin517/resources/reading/},
year={2022},
month=aug,
language={en},
author={Fruehwald, Josef}
}

@misc{eisner-2009-how,
title={How to Read a Technical Paper},
url={https://www.cs.jhu.edu/~jason/advice/how-to-read-a-paper.html},
year={2009},
language={en},
author={Eisner, Jason}
}

@inproceedings{subramonian-etal-2023-takes,
    title = "It Takes Two to Tango: Navigating Conceptualizations of {NLP} Tasks and Measurements of Performance",
    author = "Subramonian, Arjun  and
      Yuan, Xingdi  and
      Daum{\'e} III, Hal  and
      Blodgett, Su Lin",
    editor = "Rogers, Anna  and
      Boyd-Graber, Jordan  and
      Okazaki, Naoaki",
    booktitle = "Findings of the Association for Computational Linguistics: ACL 2023",
    month = jul,
    year = "2023",
    address = "Toronto, Canada",
    publisher = "Association for Computational Linguistics",
    url = "https://aclanthology.org/2023.findings-acl.202/",
    doi = "10.18653/v1/2023.findings-acl.202",
    pages = "3234--3279"
}

@article{steidl-werum-2019-operationalization,
author = {Steidl, Christina R. and Werum, Regina},
title = {If all you have is a hammer, everything looks like a nail: Operationalization matters},
journal = {Sociology Compass},
volume = {13},
number = {8},
pages = {e12727},
doi = {https://doi.org/10.1111/soc4.12727},
url = {https://compass.onlinelibrary.wiley.com/doi/abs/10.1111/soc4.12727},
eprint = {https://compass.onlinelibrary.wiley.com/doi/pdf/10.1111/soc4.12727},
note = {e12727 SOCO-1495.R2},
year = {2019}
}

@article{wood-bruner-ross-1976-role,
author = {Wood, David and Bruner, Jerome S. and Ross, Gail},
title = {THE ROLE OF TUTORING IN PROBLEM SOLVING},
journal = {Journal of Child Psychology and Psychiatry},
volume = {17},
number = {2},
pages = {89-100},
doi = {https://doi.org/10.1111/j.1469-7610.1976.tb00381.x},
url = {https://acamh.onlinelibrary.wiley.com/doi/abs/10.1111/j.1469-7610.1976.tb00381.x},
eprint = {https://acamh.onlinelibrary.wiley.com/doi/pdf/10.1111/j.1469-7610.1976.tb00381.x},
year = {1976}
}

@misc{deepseekai2025deepseekr1incentivizingreasoningcapability,
      title={DeepSeek-R1: Incentivizing Reasoning Capability in LLMs via Reinforcement Learning}, 
      author={DeepSeek-AI and Daya Guo and Dejian Yang and Haowei Zhang and Junxiao Song and Ruoyu Zhang and Runxin Xu and Qihao Zhu and Shirong Ma and Peiyi Wang and Xiao Bi and Xiaokang Zhang and Xingkai Yu and Yu Wu and Z. F. Wu and Zhibin Gou and Zhihong Shao and Zhuoshu Li and Ziyi Gao and Aixin Liu and Bing Xue and Bingxuan Wang and Bochao Wu and Bei Feng and Chengda Lu and Chenggang Zhao and Chengqi Deng and Chenyu Zhang and Chong Ruan and Damai Dai and Deli Chen and Dongjie Ji and Erhang Li and Fangyun Lin and Fucong Dai and Fuli Luo and Guangbo Hao and Guanting Chen and Guowei Li and H. Zhang and Han Bao and Hanwei Xu and Haocheng Wang and Honghui Ding and Huajian Xin and Huazuo Gao and Hui Qu and Hui Li and Jianzhong Guo and Jiashi Li and Jiawei Wang and Jingchang Chen and Jingyang Yuan and Junjie Qiu and Junlong Li and J. L. Cai and Jiaqi Ni and Jian Liang and Jin Chen and Kai Dong and Kai Hu and Kaige Gao and Kang Guan and Kexin Huang and Kuai Yu and Lean Wang and Lecong Zhang and Liang Zhao and Litong Wang and Liyue Zhang and Lei Xu and Leyi Xia and Mingchuan Zhang and Minghua Zhang and Minghui Tang and Meng Li and Miaojun Wang and Mingming Li and Ning Tian and Panpan Huang and Peng Zhang and Qiancheng Wang and Qinyu Chen and Qiushi Du and Ruiqi Ge and Ruisong Zhang and Ruizhe Pan and Runji Wang and R. J. Chen and R. L. Jin and Ruyi Chen and Shanghao Lu and Shangyan Zhou and Shanhuang Chen and Shengfeng Ye and Shiyu Wang and Shuiping Yu and Shunfeng Zhou and Shuting Pan and S. S. Li and Shuang Zhou and Shaoqing Wu and Shengfeng Ye and Tao Yun and Tian Pei and Tianyu Sun and T. Wang and Wangding Zeng and Wanjia Zhao and Wen Liu and Wenfeng Liang and Wenjun Gao and Wenqin Yu and Wentao Zhang and W. L. Xiao and Wei An and Xiaodong Liu and Xiaohan Wang and Xiaokang Chen and Xiaotao Nie and Xin Cheng and Xin Liu and Xin Xie and Xingchao Liu and Xinyu Yang and Xinyuan Li and Xuecheng Su and Xuheng Lin and X. Q. Li and Xiangyue Jin and Xiaojin Shen and Xiaosha Chen and Xiaowen Sun and Xiaoxiang Wang and Xinnan Song and Xinyi Zhou and Xianzu Wang and Xinxia Shan and Y. K. Li and Y. Q. Wang and Y. X. Wei and Yang Zhang and Yanhong Xu and Yao Li and Yao Zhao and Yaofeng Sun and Yaohui Wang and Yi Yu and Yichao Zhang and Yifan Shi and Yiliang Xiong and Ying He and Yishi Piao and Yisong Wang and Yixuan Tan and Yiyang Ma and Yiyuan Liu and Yongqiang Guo and Yuan Ou and Yuduan Wang and Yue Gong and Yuheng Zou and Yujia He and Yunfan Xiong and Yuxiang Luo and Yuxiang You and Yuxuan Liu and Yuyang Zhou and Y. X. Zhu and Yanhong Xu and Yanping Huang and Yaohui Li and Yi Zheng and Yuchen Zhu and Yunxian Ma and Ying Tang and Yukun Zha and Yuting Yan and Z. Z. Ren and Zehui Ren and Zhangli Sha and Zhe Fu and Zhean Xu and Zhenda Xie and Zhengyan Zhang and Zhewen Hao and Zhicheng Ma and Zhigang Yan and Zhiyu Wu and Zihui Gu and Zijia Zhu and Zijun Liu and Zilin Li and Ziwei Xie and Ziyang Song and Zizheng Pan and Zhen Huang and Zhipeng Xu and Zhongyu Zhang and Zhen Zhang},
      year={2025},
      eprint={2501.12948},
      archivePrefix={arXiv},
      primaryClass={cs.CL},
      url={https://arxiv.org/abs/2501.12948}, 
}

@article{lipton-2018-mythos,
author = {Lipton, Zachary C.},
title = {The mythos of model interpretability},
year = {2018},
issue_date = {October 2018},
publisher = {Association for Computing Machinery},
address = {New York, NY, USA},
volume = {61},
number = {10},
issn = {0001-0782},
url = {https://doi.org/10.1145/3233231},
doi = {10.1145/3233231},
journal = {Commun. ACM},
month = sep,
pages = {36–43},
numpages = {8}
}

@article{Krishnan2020-KRIAIA-3,
	author = {Maya Krishnan},
	doi = {10.1007/s13347-019-00372-9},
	journal = {Philosophy and Technology},
	number = {3},
	pages = {487--502},
	publisher = {Springer Verlag},
	title = {Against Interpretability: A Critical Examination of the Interpretability Problem in Machine Learning},
	volume = {33},
	year = {2020}
}

@inproceedings{
ghandeharioun2024patchscopes,
title={Patchscopes: A Unifying Framework for Inspecting Hidden Representations of Language Models},
author={Asma Ghandeharioun and Avi Caciularu and Adam Pearce and Lucas Dixon and Mor Geva},
booktitle={Forty-first International Conference on Machine Learning},
year={2024},
url={https://openreview.net/forum?id=5uwBzcn885}
}

@inproceedings{huang-etal-2024-ravel,
    title = "{RAVEL}: Evaluating Interpretability Methods on Disentangling Language Model Representations",
    author = "Huang, Jing  and
      Wu, Zhengxuan  and
      Potts, Christopher  and
      Geva, Mor  and
      Geiger, Atticus",
    editor = "Ku, Lun-Wei  and
      Martins, Andre  and
      Srikumar, Vivek",
    booktitle = "Proceedings of the 62nd Annual Meeting of the Association for Computational Linguistics (Volume 1: Long Papers)",
    month = aug,
    year = "2024",
    address = "Bangkok, Thailand",
    publisher = "Association for Computational Linguistics",
    url = "https://aclanthology.org/2024.acl-long.470/",
    doi = "10.18653/v1/2024.acl-long.470",
    pages = "8669--8687"
}

@article{MILLER20191,
title = {Explanation in artificial intelligence: Insights from the social sciences},
journal = {Artificial Intelligence},
volume = {267},
pages = {1-38},
year = {2019},
issn = {0004-3702},
doi = {https://doi.org/10.1016/j.artint.2018.07.007},
url = {https://www.sciencedirect.com/science/article/pii/S0004370218305988},
author = {Tim Miller}
}

@inproceedings{jain-wallace-2019-attention,
    title = "{A}ttention is not {E}xplanation",
    author = "Jain, Sarthak  and
      Wallace, Byron C.",
    editor = "Burstein, Jill  and
      Doran, Christy  and
      Solorio, Thamar",
    booktitle = "Proceedings of the 2019 Conference of the North {A}merican Chapter of the Association for Computational Linguistics: Human Language Technologies, Volume 1 (Long and Short Papers)",
    month = jun,
    year = "2019",
    address = "Minneapolis, Minnesota",
    publisher = "Association for Computational Linguistics",
    url = "https://aclanthology.org/N19-1357/",
    doi = "10.18653/v1/N19-1357",
    pages = "3543--3556"
}

@inproceedings{wiegreffe-pinter-2019-attention,
    title = "Attention is not not Explanation",
    author = "Wiegreffe, Sarah  and
      Pinter, Yuval",
    editor = "Inui, Kentaro  and
      Jiang, Jing  and
      Ng, Vincent  and
      Wan, Xiaojun",
    booktitle = "Proceedings of the 2019 Conference on Empirical Methods in Natural Language Processing and the 9th International Joint Conference on Natural Language Processing (EMNLP-IJCNLP)",
    month = nov,
    year = "2019",
    address = "Hong Kong, China",
    publisher = "Association for Computational Linguistics",
    url = "https://aclanthology.org/D19-1002/",
    doi = "10.18653/v1/D19-1002",
    pages = "11--20"
}

@inproceedings{bender-koller-2020-climbing,
    title = "Climbing towards {NLU}: {On} Meaning, Form, and Understanding in the Age of Data",
    author = "Bender, Emily M.  and
      Koller, Alexander",
    editor = "Jurafsky, Dan  and
      Chai, Joyce  and
      Schluter, Natalie  and
      Tetreault, Joel",
    booktitle = "Proceedings of the 58th Annual Meeting of the Association for Computational Linguistics",
    month = jul,
    year = "2020",
    address = "Online",
    publisher = "Association for Computational Linguistics",
    url = "https://aclanthology.org/2020.acl-main.463/",
    doi = "10.18653/v1/2020.acl-main.463",
    pages = "5185--5198"
}

@inproceedings{blodgett-etal-2021-stereotyping,
    title = "Stereotyping {N}orwegian Salmon: An Inventory of Pitfalls in Fairness Benchmark Datasets",
    author = "Blodgett, Su Lin  and
      Lopez, Gilsinia  and
      Olteanu, Alexandra  and
      Sim, Robert  and
      Wallach, Hanna",
    editor = "Zong, Chengqing  and
      Xia, Fei  and
      Li, Wenjie  and
      Navigli, Roberto",
    booktitle = "Proceedings of the 59th Annual Meeting of the Association for Computational Linguistics and the 11th International Joint Conference on Natural Language Processing (Volume 1: Long Papers)",
    month = aug,
    year = "2021",
    address = "Online",
    publisher = "Association for Computational Linguistics",
    url = "https://aclanthology.org/2021.acl-long.81/",
    doi = "10.18653/v1/2021.acl-long.81",
    pages = "1004--1015"
}

@misc{michael-2020-dissect,
title={To Dissect an Octopus: Making Sense of the Form/Meaning Debate },
url={https://julianmichael.org/blog/2020/07/23/to-dissect-an-octopus.html},
year={2020},
language={en},
author={Michael, Julian}
}

\appendix

\section{Full List of Readings}
\label{app:assigned-readings}

A selection of critique/conceptual papers (and possible content papers in parentheses) is presented by concept below.
Content papers may be swapped out in future iterations of the course for more recent or relevant work (or simply so that the instructor can read something new).

\paragraph{Stereotypes.}
\citet{blodgett-etal-2021-stereotyping}

\noindent\citep{ungless-etal-2023-stereotypes,mitchell-etal-2025-shades,rowe-etal-2025-eurogest}

\paragraph{Interpretability.}
\citet{lipton-2018-mythos,Krishnan2020-KRIAIA-3,saphra-wiegreffe-2024-mechanistic}

\noindent\citep{ghandeharioun2024patchscopes,huang-etal-2024-ravel}

\paragraph{Explainability.}
\citet{miller2017explainableaibewareinmates,MILLER20191,bastings-filippova-2020-elephant}

\noindent\citep{jain-wallace-2019-attention,wiegreffe-pinter-2019-attention,jain-etal-2020-learning}

\paragraph{Paraphrases.}
\citet{bhagat-hovy-2013-squibs}

\noindent\citep{wahle-etal-2022-large,wahle-etal-2023-paraphrase,sharma-etal-2023-paraphrase}

\paragraph{Bias.}
\citet{blodgett-etal-2020-language}

\noindent\citep{sap-etal-2020-social,goldfarb-tarrant-etal-2021-intrinsic,sap-etal-2022-annotators,parmar-etal-2023-dont}

\paragraph{Race.}
\citet{field-etal-2021-survey}

\citep{sap-etal-2019-risk,deas-etal-2023-evaluation,bourgeade-etal-2023-multilingual}

\paragraph{Emergent abilities.}
\citet{schaeffer2023emergentabilitieslargelanguage}

\citep{wei2022emergent,lu-etal-2024-emergent,liu-etal-2024-emergent}

\paragraph{Gender.}
\citet{larson-2017-gender,devinney-et-al-2022-theories}

\citep{vashishtha-etal-2023-evaluating,waldis-etal-2024-lou,gautam-etal-2024-winopron}

\paragraph{Generalization.}
\citet{Hupkes_Giulianelli_Dankers_Artetxe_Elazar_Pimentel_Christodoulopoulos_Lasri_Saphra_Sinclair_etal}

\citep{weissweiler-etal-2023-counting,muennighoff-etal-2023-crosslingual,ross-etal-2024-artificial}

\paragraph{Names.}
\citet{gautam-etal-2024-stop}

\citep{Asr_Mazraeh_Lopes_Gautam_Gonzales_Rao_Taboada_2021,sandoval-etal-2023-rose,an-rudinger-2023-nichelle}

\paragraph{Reasoning.}
\citet{kargupta2025cognitivefoundationsreasoningmanifestation}

\citep{wu-etal-2024-reasoning,gautam-et-al-2024-robust,deepseekai2025deepseekr1incentivizingreasoningcapability}

\section{Course Requirements and Instructions}
\label{app:course-requirements}

For 7 credits, you are required to do a presentation and write a final report. You will be graded on: Engagement (30\%), presenting a concept and leading a discussion (35\%), and your final report (35\%).

For 4 credits, you are only required to do a presentation. You will be graded on: Engagement (40\%), and presenting a concept and leading a discussion (60\%).

\paragraph{Assignments.}

\begin{itemize}
    \item Read the critique paper(s) first
    \item Then read (or do a quick scan, if you're short on time) the content papers
    \item Write 3-5 bullet points of synthesis/critique (you may also write more if you're feeling particularly inspired)
        \begin{itemize}
            \item Connect ideas / definitions / methods across different papers
            \item Critique what's missing or suspicious about the content papers using what you learned from the critique papers
            \item Name some arguments that a paper makes and tell me whether you are convinced by them or not, and if so, why
        \end{itemize}
    \item For maximum points, I want to see that you engaged with all 4 of the papers beyond repeating their content
    \item \textbf{Do not use LLMs}: Please respect the time and effort I put it into reading your responses and giving you feedback, by putting your own thoughts down
\end{itemize}

\paragraph{Engagement.}
\begin{itemize}
    \item Mixture of assignments and in-class participation
    \item Minimum 33\% each required to pass
    \item Assignments are to encourage you to do the readings and come to class prepared, because this is a discussion-heavy class
    \item In-class participation includes asking questions, answering questions and discussion prompts, etc.
\end{itemize}

\paragraph{Presentation + leading a discussion.}
\begin{itemize}
    \item Timing: 1 hour - 1 hour 15 mins
    \item Content should synthesize the readings
        \begin{itemize}
            \item Look for commonalities and differences across papers (including from other concepts that we previously covered)
            \item Discuss conceptualization and operationalization of your concept, and how they differ/coincide across the content papers
            \item Critically question all assumptions that papers make
            \item Try to answer the question ``\textit{Why?}'' Why are we doing this work in the field? For what purpose is it useful? etc.
        \end{itemize}
    \item Don't present the entire time
        \begin{itemize}
            \item Your audience has also read the papers
            \item Engage them meaningfully at \textbf{least} every 10-15 minutes (more is fine!)
            \item I encourage the use of questions and discussion prompts for this
            \item Ensure that everyone gets a chance to speak
        \end{itemize}
    \item Meta stuff (also graded!)
        \begin{itemize}
            \item Have clear / visually appealing slides (not too crowded / text-heavy)
            \item Show up early to set up
            \item Message me in advance to remind me if your team will need special cables or if you will be presenting from my laptop
        \end{itemize}
\end{itemize}

\paragraph{Report.}
\begin{itemize}
    \item Pick a concept (does not have to be the one you presented on) and design a new research project that addresses a gap in how it is currently conceptualized / operationalized according to critiques (you can also re-design one of the content papers we read instead of making a new one, if you want)
    \item Your title should explicitly state the concept you picked
    \item 4 pages not including references, 11 or 12 point font, ACL ARR template
    \item Structure
        \begin{itemize}
            \item Abstract
            \item Introduction (motivate the gap in the literature that you want to fill, i.e., explain what the gap is in conceptualization/operationalization, and why it matters)
            \item Experimental methodology (explain how, concretely, you will address the problem - If you will propose a new dataset, how will you design it? If you are proposing a new method, how will you evaluate it? If you’re doing a new evaluation, what is the evaluation metric you are going to propose/use and why? If you propose all of the above I expect all parts to be well-motivated)
            \item Related work (search for 10 or more papers on your concept that are closely related, for which you will do a quick scan and write a sentence each about how your work relates to and differs from theirs; this is to contextualize your work in the broader landscape of NLP research)
            \item Expected impact (explain how you expect your work to benefit the research community / advance science!)
            \item Conclusion
        \end{itemize}
        \item \textbf{No AI, please!} I want to see your raw thoughts and writing so you can learn how to structure your thoughts in written form and so I can give you feedback for later, and I do not want to see references to papers that don’t exist
\end{itemize}

\end{document}